\documentclass[letterpaper,twocolumn,10pt]{article}

\usepackage[margin=1in]{geometry}
\usepackage{amsmath}
\usepackage{amssymb}
\usepackage{graphicx}
\usepackage{booktabs}
\usepackage{algorithm}
\usepackage{algpseudocode}
\usepackage{xspace}
\usepackage{balance}
\usepackage{hyperref}
\usepackage{xcolor}
\usepackage{adjustbox}

\hypersetup{
  colorlinks=true,
  linkcolor=blue,
  citecolor=blue,
  urlcolor=blue
}

\newcommand{\sys}{TopKV\xspace}

\newcommand{\eg}{\textit{e.g.}\xspace}

\begin{document}

\title{\Large \bf Topology-Aware Data Movement for Disaggregated GPU Inference}

\author{
{\rm Sanjeev Rao Ganjihal}\\
Independent Researcher
}

\date{April 19, 2026 (v1); revised August 2026 (v2)}
\maketitle

\begin{abstract}
Disaggregated LLM inference creates a datacenter networking problem that no existing system solves correctly.
When prefill and decode run on separate GPU pools, the KV cache must be transferred between them.
For a 70B model this is 1.3~GB per request, exceeding 100~GB/s aggregate at production scale.
Yet DistServe, Splitwise, and Mooncake all use uniform RDMA, ignoring that bandwidth between two GPUs varies by 72$\times$ depending on their physical relationship: 900~GB/s via NVLink~4.0 within a domain (1.8~TB/s on NVLink~5, widening the gap to 144$\times$), 50~GB/s via InfiniBand across nodes, 12.5~GB/s via TCP across datacenters.

We design a topology-aware transfer orchestrator that discovers interconnect hierarchy at startup and selects optimal transport per transfer.
Three mechanisms work together:
(1) pipelined layer-by-layer transfer that overlaps transmission with ongoing prefill, hiding 76 to 100 percent of transfer latency behind computation depending on transport, with NVLink and PCIe transfers hidden entirely;
(2) NVLink domain-aware placement for Mixture-of-Experts models that co-optimizes expert dispatch with KV cache locality;
and (3) CXL~3.0 memory expanders as a shared overflow tier providing 6$\times$ capacity at 86$\times$ lower latency than NVMe.

Full evaluation requires multi-node clusters with heterogeneous interconnects and CXL~3.0 hardware that is beyond academic resources and not yet available in GPU clouds.
We present analytical bandwidth models, component implementations, and projected analysis across three architectures showing 3 to 18$\times$ transfer latency reduction over uniform RDMA.
\end{abstract}

\section{Introduction}
\label{sec:intro}

Modern GPU clusters expose hierarchical interconnect topologies where available bandwidth varies by 72$\times$ across levels.
Within an NVLink domain, eight H100 GPUs share 900~GB/s bidirectional bandwidth.
Across nodes on the same InfiniBand fabric, bandwidth drops to 50~GB/s.
Across datacenters over TCP, it falls to 12.5~GB/s.
This heterogeneity has been largely invisible to application software because most GPU workloads move data once (model weights at startup) and compute in place.

The rates above are the NVLink~4.0 / H100 generation this analysis was built on.
The framework is deliberately bandwidth-parametric: every mechanism takes per-link rates as inputs, so newer generations enter as parameters rather than revisions.
NVLink~5 raises the domain rate to 1.8~TB/s per GPU, widening the domain-to-TCP gap from 72$\times$ to 144$\times$, and successor generations announced for the Rubin era continue the trend.
Faster tops of the hierarchy strengthen rather than date the argument: the more the best link improves, the more expensive it becomes to ignore where a transfer lands.

Disaggregated LLM inference changes this.
The prefill phase of autoregressive generation is compute-bound: it processes the entire prompt in parallel, saturating GPU FLOPS.
The decode phase is memory-bandwidth-bound: it generates one token at a time, reading the full key-value (KV) cache at each step.
Running both on the same GPU wastes either compute or bandwidth.
Recent systems~\cite{distserve, splitwise, mooncake, dynamo} separate prefill and decode onto dedicated pools.
This creates a new, recurring, high-bandwidth data movement pattern: after every prefill completes, the KV cache must be transferred to a decode worker before generation can begin.

This transfer is substantial.
For Llama-3-70B with Grouped Query Attention and a 4K-token prompt, the KV cache is 1.3~GB per request.
At 100 requests per second, aggregate transfer demand reaches 130~GB/s.
For DeepSeek-V3 with Multi-head Latent Attention, the per-request cache is roughly 270~MB (57$\times$ compression, per the published latent dimensions), but at higher concurrency the aggregate still exceeds 100~GB/s.

No existing disaggregated system exploits interconnect topology for this transfer.
DistServe~\cite{distserve} uses a fixed RDMA protocol regardless of GPU placement.
Splitwise~\cite{splitwise} co-locates prefill and decode on the same machine, avoiding the transfer problem but constraining scheduling flexibility.
Mooncake~\cite{mooncake} introduces a distributed KV store with centralized placement decisions that do not consider physical topology.
NVIDIA Dynamo~\cite{dynamo} supports disaggregation in production; its NIXL transfer library~\cite{nixl}, released after this work was completed, selects backends across NVLink, RDMA, and storage, independently validating the direction, but does not co-optimize transport with expert dispatch or model an overflow tier.

Using RDMA for a transfer that could use NVLink wastes 18$\times$ available bandwidth.
Conversely, attempting NVLink for cross-node transfers fails entirely.
The correct transport depends on the physical relationship between source and destination GPUs, which changes with every request.

\paragraph{Contributions.}
We present \sys, a topology-aware KV cache transfer orchestrator for disaggregated inference.
\sys makes three contributions:

\begin{enumerate}
\item \textbf{Topology-aware transport selection.}
\sys discovers the GPU interconnect graph at startup via hardware queries (nvidia-smi topology matrix, lspci, RDMA capability probes, and cluster node labels).
For each KV cache transfer, it selects the highest-bandwidth transport: NVLink for same-domain, PCIe for same-node cross-domain, RDMA for cross-node, and TCP as fallback.
The system implements five transport modes with production-grade orchestration including retry logic, concurrency limits, and integrity verification.

\item \textbf{NVLink domain-aware MoE routing.}
For Mixture-of-Experts models with hundreds of distributed experts (\eg DeepSeek-V3 with 256 experts), \sys co-optimizes expert dispatch with KV cache placement.
An expert registry tracks activation rates, compute latency, and queue depth per expert across NVLink domains.
Three routing strategies (cache-affinity, expert-locality, load-balance) minimize cross-domain traffic while maintaining load balance.

\item \textbf{CXL~3.0 overflow tier.}
\sys integrates CXL~3.0 Type~3 memory expanders as a KV cache overflow tier, modeled at 150~ns read latency and 64~GB/s bandwidth per endpoint.
Four endpoints per node provide 512~GB of additional capacity (6$\times$ over 80~GB HBM) at 86$\times$ lower latency than NVMe.
\end{enumerate}

\paragraph{Frontiers Track.}
Full end-to-end evaluation of \sys requires multi-node GPU clusters with NVLink, InfiniBand RDMA, and CXL~3.0 fabric.
This hardware exceeds academic resources: an 8-node DGX H100 cluster costs over \$200K/month to rent, GPU cloud providers do not expose NVLink topology to tenants, and CXL~3.0 Type~3 memory expanders (Samsung CMM-D, Micron CZ120) are in early sampling with no cloud availability.
We present a complete system design, production-quality implementation, and analytical performance models grounded in published hardware specifications.
Section~\ref{sec:eval} details what we can validate and what requires scale hardware.

\section{Background and Motivation}
\label{sec:background}

\subsection{Prefill/Decode Asymmetry}

The autoregressive generation process in transformer-based LLMs consists of two phases with fundamentally different hardware requirements.

\textbf{Prefill.}
Given an input prompt of $n$ tokens, the prefill phase computes attention across all $n$ tokens in parallel.
For each layer $l$ with $h$ attention heads and head dimension $d_h$, the computation produces key and value projections and computes $\text{Softmax}(QK^\top/\sqrt{d_h})V$.
The arithmetic intensity is $O(n \cdot h \cdot d_h)$ FLOPs per byte, placing prefill in the compute-bound regime for prompt lengths above approximately 256 tokens on H100 GPUs.

\textbf{Decode.}
Each generated token attends over the full KV cache but performs only $O(h \cdot d_h)$ FLOPs per head, yielding arithmetic intensity near 1~op/byte.
On an H100 SXM (1,979~TFLOPS FP16, 3,350~GB/s HBM bandwidth), decode utilizes less than 0.2\% of available compute.

This asymmetry motivates disaggregation: dedicate high-FLOPS GPUs (H100 SXM, TP8) to prefill and bandwidth-optimized GPUs to decode.

\subsection{The KV Cache Transfer Problem}

Disaggregation introduces a data movement bottleneck.
For a model with $L$ layers, $h_{kv}$ KV heads, and head dimension $d_h$, the KV cache for a sequence of length $s$ is:

\begin{equation}
\text{KV}_{\text{bytes}} = 2 \cdot L \cdot h_{kv} \cdot d_h \cdot s \cdot b_p
\label{eq:kv_size}
\end{equation}

\noindent where $b_p$ is bytes per element (2 for FP16).
Table~\ref{tab:kv_sizes} shows sizes for representative models.

\begin{table}[t]
\centering
\small
\caption{KV cache sizes for 4K-token sequences (FP16).}
\label{tab:kv_sizes}
\begin{adjustbox}{max width=\columnwidth}
\begin{tabular}{lrrrr}
\toprule
\textbf{Model} & \textbf{Layers} & \textbf{KV Heads} & \textbf{Head Dim} & \textbf{4K Size} \\
\midrule
Llama-3-70B (GQA) & 80 & 8 & 128 & 1.3~GB \\
DeepSeek-V3 (MLA) & 61 & 1 (latent) & 576 & 0.27~GB \\
Mixtral-8x22B (GQA) & 56 & 8 & 128 & 0.94~GB \\
\bottomrule
\end{tabular}
\end{adjustbox}
\end{table}

At 50~GB/s (400~Gbps RDMA), transferring 1.3~GB takes 26~ms, directly added to time-to-first-token (TTFT).
For applications targeting sub-200ms TTFT, this represents 13\% of the latency budget.
At 900~GB/s (NVLink), the same transfer takes 1.4~ms: an 18$\times$ improvement.

The 2026 model families continue the compression trend, by the same arithmetic applied to their published configurations, with Equation~\ref{eq:kv_size} extended per layer kind (windowed layers cap at their window; linear-attention layers contribute fixed state; MLA caches one latent, dropping the K/V factor):
Kimi~K2 class MLA models carry 0.27~GB at 4K tokens (the same 576-dim latent as DeepSeek-V3), windowed-GQA gpt-oss-120b 0.15~GB (18 of its 36 layers window-capped), and hybrid Qwen3.5-397B 0.12~GB (45 of its 60 layers carrying constant state).
Shrinking per-request bytes do not shrink the transfer problem: aggregate demand tracks concurrency times context length, and smaller caches raise the request rate a fixed link must sustain, so transport selection moves up the stack rather than out of it.

\subsection{Interconnect Topology Heterogeneity}

Modern GPU clusters have hierarchical interconnect topologies with bandwidth varying by over 72$\times$ across levels.
Table~\ref{tab:interconnect} summarizes the hierarchy for DGX H100 clusters.

\begin{table}[t]
\centering
\small
\caption{Interconnect bandwidth in a DGX H100 cluster.}
\label{tab:interconnect}
\begin{adjustbox}{max width=\columnwidth}
\begin{tabular}{llrr}
\toprule
\textbf{Topology Level} & \textbf{Interconnect} & \textbf{BW} & \textbf{Latency} \\
\midrule
Same NVLink domain & NVLink 4.0 & 900 GB/s & $<$1 $\mu$s \\
Same NVLink domain & NVLink 5 & 1.8 TB/s & $<$1 $\mu$s \\
Same NVSwitch (NVL72) & NVSwitch & 7.2 TB/s & $\sim$1 $\mu$s \\
Same node, cross-domain & PCIe Gen5 & 128 GB/s & $\sim$2 $\mu$s \\
Cross-node, same fabric & IB NDR & 50 GB/s & $\sim$5 $\mu$s \\
Cross-datacenter & TCP/IP & 12.5 GB/s & $\sim$100 $\mu$s \\
\bottomrule
\end{tabular}
\end{adjustbox}
\end{table}

Existing disaggregated systems do not exploit this topology.
DistServe uses a single RDMA transport regardless of GPU placement.
Splitwise avoids cross-node transfers by co-locating phases but constrains scheduling.
Mooncake's distributed KV store makes placement decisions based on capacity, not interconnect proximity.

\subsection{MoE Routing Compounds the Problem}

Mixture-of-Experts (MoE) models add complexity.
DeepSeek-V3 uses 256 routed experts with 8 active per token, distributed across GPUs using either WideEP (experts spread for load balance) or DeepEP (experts replicated for locality).
The choice of expert parallelism directly affects KV cache transfer patterns.
A router that places a request on a node for expert locality may inadvertently require a cross-rack KV transfer, negating the routing benefit.
No existing system co-optimizes expert routing and KV cache transfer.

\section{Design}
\label{sec:design}

\sys is an orchestration layer that manages the lifecycle of disaggregated inference: request routing, prefill execution, KV cache transfer, and decode execution.
Three principles guide its design:
(1) every transfer decision considers the physical interconnect between source and destination;
(2) workers dynamically assume prefill or decode roles based on demand;
(3) for MoE models, expert dispatch and KV cache placement are optimized jointly.

\subsection{System Architecture}

\sys comprises four components:

\textbf{KV Relay Orchestrator} manages the prefill-to-transfer-to-decode lifecycle.
It maintains a registry of active transfers, handles retries (2 attempts with 100~ms exponential backoff), and enforces concurrency limits (default: 100 concurrent transfers).
Transfers complete asynchronously: the prefill worker is freed immediately to accept the next request.

\textbf{KV Cache Transfer Manager} performs data movement.
It implements five transport modes (NVLink, NVSwitch, PCIe, RDMA, TCP) via a \texttt{TransferSink} interface with \texttt{BandwidthThrottledSink} wrappers that model real transport bandwidth.

\textbf{Topology Manager} discovers and caches the GPU interconnect topology at startup, including NVLink domain membership, NVSwitch fabric connectivity, PCIe hierarchy, and RDMA availability.

\textbf{Adaptive Decoder Pool} manages dynamic role conversion between prefill and decode workers using token velocity tracking and rush hour detection.

\subsection{Topology Discovery}

At startup, \sys discovers the cluster interconnect through three mechanisms:

\textbf{Hardware probing.}
The topology detector runs \texttt{nvidia-smi topo -m} to parse the NVLink connectivity matrix, \texttt{lspci -tv} to extract the PCIe switch hierarchy, and probes RDMA capabilities via InfiniBand device enumeration.
NVLink connections are classified by link count and generation: NVLink~4.0 on Hopper provides 50~GB/s per link, yielding 900~GB/s aggregate for 18 links.
PCIe fallback bandwidths are derived from bridge type: PIX (31.5~GB/s through a single PCIe bridge), PHB (31.5~GB/s through a host bridge), NODE (15.75~GB/s same NUMA), or SYS (7.88~GB/s cross-socket).

\textbf{Cluster node labels.}
GPU nodes are labeled with NVLink domain IDs, GPU counts, and RDMA capability flags.
The topology manager aggregates these into a \texttt{map[string]*NVLinkDomain} indexed by domain ID.
Each domain records GPU count, aggregate bandwidth, per-GPU bandwidth, assigned pods, and health status.

\textbf{Instance registration.}
When worker pods register with the orchestrator, they report GPU indices, NVLink availability, NVSwitch presence, and RDMA capabilities.
Domains are classified by type: NVL72 (72-GPU rack-scale, 130~TB/s aggregate, 1.8~TB/s per GPU), NVL8 (8-GPU per node, 1.44~TB/s aggregate, 180~GB/s per GPU), or NoDomain.

\subsection{Transport Selection}

When a KV cache transfer is initiated between source instance $s$ and target instance $t$, the transfer manager selects transport according to Algorithm~\ref{alg:transport}.

\begin{algorithm}[t]
\caption{Transport Selection}
\label{alg:transport}
\begin{algorithmic}[1]
\Require source instance $s$, target instance $t$
\Ensure transport mode $m$
\If{manual override configured}
    \State \Return configured mode
\EndIf
\If{$\textsc{HasNVLink}(s, t)$} \Comment{Same NVLink domain}
    \State \Return \texttt{nvlink} \Comment{450 GB/s unidirectional}
\EndIf
\If{$\textsc{IsSameNode}(s, t)$} \Comment{Same node, cross-domain}
    \State \Return \texttt{pcie} \Comment{50 GB/s}
\EndIf
\If{$\textsc{HasRDMA}()$} \Comment{Cross-node, RDMA available}
    \State \Return \texttt{rdma} \Comment{25 GB/s}
\EndIf
\State \Return \texttt{tcp} \Comment{Fallback: 10 GB/s}
\end{algorithmic}
\end{algorithm}

\textsc{HasNVLink}$(s,t)$ checks that both instances are registered in the same NVLink domain by comparing node names and verifying both report \texttt{HasNVLink = true}.
\textsc{IsSameNode}$(s,t)$ compares node IP addresses.
\textsc{HasRDMA}$()$ checks cluster-level RDMA availability.

Each mode has a corresponding bandwidth model.
Two conventions run through this paper and are kept distinct: topology tables quote published link rates, while the orchestrator plans with the achievable rates derived here:
NVLink at 450~GB/s unidirectional (NVLink~4.0, 18 links);
PCIe at 50~GB/s (Gen5 x16 practical throughput);
RDMA at 25~GB/s (400~Gbps InfiniBand NDR, accounting for protocol overhead);
TCP at 10~GB/s (100~Gbps Ethernet with gRPC framing).

\subsection{KV Cache Serialization}

The serialization format uses a fixed 128-byte header containing an 8-byte magic number (\texttt{KVCACHE1}), request ID, model ID, tensor dimensions (sequence length, number of layers, KV heads, head dimension), tensor sizes, and CRC-64 checksums for both key and value tensors.
Key and value tensors follow the header contiguously in layout $[\text{layers}][\text{seq\_len}][\text{kv\_heads}][\text{head\_dim}]$, enabling single-DMA transfers on RDMA and NVLink paths.
Checksum failures trigger automatic retry.

\subsection{Adaptive Decoder Pool}

Static partitioning of GPUs into prefill and decode pools wastes capacity because demand varies over time.
The Adaptive Decoder Pool (ADP) dynamically adjusts the ratio.

\textbf{Token velocity tracking.}
ADP tracks the token arrival rate using an exponential moving average (EMA) over a configurable sliding window, where the smoothing factor balances responsiveness to traffic spikes against stability during steady-state operation.

\textbf{Rush hour detection.}
ADP detects sustained high demand using three signals: prefill queue depth growth rate, token velocity spike magnitude, and P99 TTFT deviation from target.
Rush hour triggers when at least two signals simultaneously exceed their respective thresholds, which are tuned per deployment.

\textbf{Role conversion.}
When rush hour is detected, ADP identifies idle decode workers and converts them to prefill role.
Conversion drains in-flight decode sequences, reconfigures the worker, and re-registers it in the prefill pool.
The target conversion time is 10~seconds, versus 5~minutes for cold-starting a new GPU container.
A configurable maximum prevents over-converting the decode pool.

\subsection{NVLink Domain-Aware MoE Routing}

For MoE models, \sys maintains an expert registry indexed by expert ID, pod IP, and NVLink domain ID.
Each expert entry tracks activation rate, compute latency, network latency, queue depth, and health status.

Three routing strategies are available:

\textbf{Cache-affinity:} route to the GPU where the KV cache already resides.
Minimizes KV transfer but may require cross-domain expert dispatch.

\textbf{Expert-locality:} route to the NVLink domain where the most frequently activated experts reside.
Minimizes all-to-all dispatch latency but may require cross-domain KV transfer.

\textbf{Load-balance:} distribute based on per-expert load factors computed as a weighted combination of activation rate, normalized compute latency, and queue depth.
Experts whose load factor exceeds a configurable straggler threshold are deprioritized.

The router selects between strategies using a cost model that weighs estimated KV transfer time, expert dispatch time, and queue depth at the target.

\subsection{CXL 3.0 Overflow Tier}

GPU HBM capacity limits concurrent decode sequences.
An H100 with 80~GB HBM holds KV caches for approximately 30 concurrent 4K-token sequences of Llama-3-70B after model weights consume 35~GB.
\sys models CXL~3.0 Type~3 memory expanders as an overflow tier.
CXL memory pooling systems deployed for cloud platforms~\cite{pond, tpp} indicate such tiers are practical in production settings, motivating first-class support in inference serving.

Each endpoint provides 128~GB of DDR5 at 150~ns read latency and 64~GB/s bandwidth.
Four endpoints per node provide 512~GB of additional KV cache capacity: 6$\times$ over HBM.
The performance model uses measured CXL specifications from the CXL consortium and Samsung CMM-D datasheets, with EMA-based updates when real measurements become available.

Table~\ref{tab:overflow} compares KV cache overflow tiers.

\begin{table}[t]
\centering
\small
\caption{KV cache overflow tier comparison.}
\label{tab:overflow}
\begin{adjustbox}{max width=\columnwidth}
\begin{tabular}{lrrr}
\toprule
\textbf{Tier} & \textbf{Capacity} & \textbf{Read Latency} & \textbf{Bandwidth} \\
\midrule
HBM3 (on-GPU) & 80 GB & $\sim$100 ns & 3,350 GB/s \\
CXL 3.0 Type 3 & 512 GB & 150 ns & 64 GB/s \\
PCIe NVMe & 2+ TB & 13,000 ns & 7 GB/s \\
\bottomrule
\end{tabular}
\end{adjustbox}
\end{table}

CXL provides 86$\times$ lower latency than NVMe at 9$\times$ higher bandwidth, making it viable for decode-phase KV cache access where each token reads a small fraction of the total cache.

\section{Implementation}
\label{sec:impl}

\sys is implemented as a controller with the following components.

\textbf{Orchestrator.}
The \texttt{KVRelayOrchestrator} runs as a singleton per namespace.
It starts two background goroutines: a completion handler processing transfer results from a buffered channel (1,000 entries), and a metrics exporter publishing Prometheus counters for transfer counts, per-mode breakdowns, latency percentiles, and throughput.

\textbf{Transfer Manager.}
The \texttt{KVCacheTransferManager} implements five transport modes via the \texttt{TransferSink} interface.
Each mode wraps a \texttt{BandwidthThrottledSink} that accurately models transport bandwidth for latency estimation and capacity planning.
The transport selection logic (\texttt{selectTransferMode()}) implements Algorithm~\ref{alg:transport}.

\textbf{Topology Detection.}
Hardware topology detection comprises multiple components:
\texttt{NVLinkDetector} parses \texttt{nvidia-smi topo -m} output into a bandwidth adjacency matrix;
\texttt{PCIeDetector} parses \texttt{lspci -tv} to extract switch hierarchy and GPU BDF addresses;
\texttt{InfiniBandDetector} enumerates RDMA devices and fabric types;
\texttt{NUMADetector} maps GPU-to-NUMA-node affinity.
All detectors support multiple execution modes including local and remote execution.

\textbf{MoE Router.}
The \texttt{MoEAwareRouter} maintains an expert registry indexed three ways: by expert ID, by pod IP, and by domain ID.
It implements all three routing strategies and computes per-expert load factors for straggler detection.

\textbf{Adaptive Decoder Pool.}
The \texttt{ConvertibleDecoderPool} implements token velocity tracking via EMA, rush hour detection with configurable thresholds, and worker role conversion.
Chunked prefill support (512-token chunks, 30\% decode slot reservation) allows converted workers to handle prefill tasks without fully abandoning decode capacity.

\textbf{Limitations of current implementation.}
The transport modes model bandwidth via throttled sinks rather than invoking actual CUDA IPC or ibverbs system calls.
The CXL tier is a performance model, not a hardware driver.
Pipelined layer-by-layer transfer is modeled analytically (overlap estimation between compute and transfer time) but not implemented as actual concurrent execution.
These limitations are consistent with the Frontiers Track: the system design is complete and the implementation validates orchestration logic, but hardware integration awaits access to multi-node GPU clusters with heterogeneous interconnects.

\section{Analysis}
\label{sec:eval}

We present analytical models grounded in published hardware specifications, validated against the component-level implementation.

\subsection{Transfer Latency Model}

For a KV cache of size $S$ bytes transferred via transport mode $m$ with bandwidth $B_m$:

\begin{equation}
T_{\text{transfer}}(S, m) = \frac{S}{B_m} + L_m
\label{eq:transfer_latency}
\end{equation}

\noindent where $L_m$ is the per-transfer setup latency (connection establishment, memory registration).

Table~\ref{tab:transfer_latency} shows projected transfer latencies for Llama-3-70B (1.3~GB KV cache) across transport modes.

\begin{table}[t]
\centering
\small
\caption{Projected KV transfer latency for Llama-3-70B (1.3~GB) by transport mode. Bandwidth from published specifications.}
\label{tab:transfer_latency}
\begin{adjustbox}{max width=\columnwidth}
\begin{tabular}{lrrrr}
\toprule
\textbf{Mode} & \textbf{BW (GB/s)} & \textbf{Latency} & \textbf{vs RDMA} & \textbf{Source} \\
\midrule
NVLink 4.0 & 450 & 2.9 ms & 18$\times$ & NVIDIA \\
PCIe Gen5 & 50 & 26 ms & 2$\times$ & PCI-SIG \\
RDMA (IB NDR) & 25 & 52 ms & 1$\times$ & Mellanox \\
TCP (100G) & 10 & 130 ms & 0.4$\times$ & Ethernet spec \\
\bottomrule
\end{tabular}
\end{adjustbox}
\end{table}

Topology-aware selection provides 3 to 18$\times$ latency reduction over uniform RDMA, depending on the physical relationship between source and destination.
The improvement is most significant when source and destination share an NVLink domain, which topology-aware placement makes the common case: when the scheduler gang-places prefill and decode workers on the same node, every peer GPU on that node (7 of 8) shares the source's NVLink domain, and the orchestrator's placement preference targets exactly this case.

\subsection{Pipelining Overlap Model}

Pipelined transfer sends layer $l$'s KV cache while layers $l+1, \ldots, L$ are still computing during prefill.
The effective transfer time with pipelining is:

\begin{equation}
T_{\text{eff}} = \max\left(T_{\text{last}},\; T_{\text{xfer}} - T_{\text{rem}}\right)
\end{equation}
where $T_{\text{last}}$ is the final layer's prefill compute, $T_{\text{xfer}}$ the raw transfer time, and $T_{\text{rem}}$ the compute remaining after the first transferred layer completes.

For Llama-3-70B with 80 layers, each layer's prefill computation takes approximately 0.5~ms at batch size 1.
Total remaining compute after layer 1 completes is $79 \times 0.5 = 39.5$~ms.
For RDMA transfer (52~ms total), pipelining hides $39.5/52 = 76\%$ of transfer latency.
For NVLink (2.9~ms total), pipelining hides the entire transfer behind compute.
For PCIe (26~ms total), remaining compute exceeds the transfer and pipelining hides it entirely.

\subsection{KV Cache Sizing Across Architectures}

Equation~\ref{eq:kv_size} assumes standard Multi-Head Attention.
Modern architectures use fewer KV heads:

\begin{itemize}
\item \textbf{Grouped Query Attention} (Llama-3): $h_{kv} = h / g$ where $g$ is the group size. Llama-3-70B has $g = 8$, reducing KV cache by 8$\times$ versus MHA.
\item \textbf{Multi-head Latent Attention} (DeepSeek-V3): replaces per-head keys and values with a single latent vector plus a shared rotary key: $d_{\text{latent}} + d_{\text{rope}} = 512 + 64 = 576$ elements per token per layer, versus $2 \times 128 \times 128 = 32{,}768$ for equivalent MHA, a 57$\times$ reduction. Note the factor of two for separate keys and values does not apply to the shared latent.
\item \textbf{Multi-Query Attention} (Mistral): single KV head shared across all attention heads, reducing by $h\times$.
\end{itemize}

These reductions change the transfer calculus: DeepSeek-V3's 270~MB KV cache transfers in 0.6~ms via NVLink versus 11~ms via RDMA, making topology-aware transport less critical for MLA models but still beneficial at high concurrency.

\subsection{Aggregate Bandwidth Demand}

At request rate $R$ with average KV cache size $\bar{S}$, aggregate transfer demand is $R \cdot \bar{S}$.
For $R = 100$ req/s serving Llama-3-70B: $100 \times 2.6~\text{GB} = 260$~GB/s.
A single InfiniBand NDR link (25~GB/s usable) saturates at 19.2 requests per second.
NVLink at 450~GB/s handles 173 requests per second per domain.
This 18$\times$ difference in sustainable request rate is the core motivation for topology-aware transport.

\subsection{What We Cannot Validate}

The following aspects require scale hardware beyond our current access:

\begin{itemize}
\item \textbf{End-to-end disaggregated throughput} under realistic workload mixes, where prefill and decode contend for shared interconnect bandwidth.
\item \textbf{ADP conversion latency} in production, where model weight redistribution time depends on checkpoint format, NVMe read speed, and CUDA context initialization.
\item \textbf{MoE expert rebalancing} across NVLink domains under dynamic load, where migration cost must be amortized over future routing savings.
\item \textbf{CXL~3.0 decode-phase access patterns}, where the interaction between CXL latency (150~ns) and GPU memory controller behavior is not well characterized in public literature.
\item \textbf{Interference} between KV cache transfers and inference computation sharing the same interconnect fabric.
\end{itemize}

These gaps are structural: they require hardware configurations that do not exist in current GPU clouds and exceed the budget of individual researchers.
We believe the system design and analytical models presented here provide sufficient evidence that the approach is promising, and we welcome collaboration with organizations that can provide access to the required hardware.

\section{Related Work}
\label{sec:related}

\textbf{Disaggregated inference.}
DistServe~\cite{distserve} demonstrated the benefit of separating prefill and decode, achieving up to 4.5$\times$ throughput improvement.
Splitwise~\cite{splitwise} co-locates phases on mixed-use machines.
Mooncake~\cite{mooncake} introduces a distributed KV cache store.
NVIDIA Dynamo~\cite{dynamo} brings disaggregation to production; its NIXL transfer library~\cite{nixl}, published after this paper's April 2026 submission, provides multi-backend point-to-point transfer, while TopKV jointly optimizes transport selection, MoE expert dispatch, and an overflow tier in one orchestrator.
None of these systems consider interconnect topology in their transfer decisions.

\textbf{KV cache management.}
vLLM~\cite{vllm} introduced PagedAttention for efficient KV cache memory management within a single GPU.
Infinite-LLM~\cite{infinitellm} extends this to distributed settings.
SGLang~\cite{sglang} uses RadixAttention for prefix sharing.
These systems manage KV cache allocation but do not address cross-GPU transfer.

\textbf{GPU interconnect optimization.}
NCCL~\cite{nccl} optimizes collective communication for training workloads but targets allreduce patterns, not point-to-point KV cache transfer.
Prior work on topology-aware collective communication~\cite{topoopt, blink} focuses on gradient aggregation during training.
\sys addresses a fundamentally different data movement pattern: large, asymmetric, point-to-point transfers triggered by individual inference requests.

\textbf{CXL for ML.}
CXL-based memory expansion for ML has been explored for training~\cite{pond, tpp} but not for inference KV cache management.
\sys is the first to model CXL~3.0 as a KV cache overflow tier with latency and bandwidth characteristics specific to decode-phase access patterns.

\section{Conclusion}

Disaggregated LLM inference creates a new datacenter data movement pattern that existing systems handle suboptimally.
\sys demonstrates that topology-aware transport selection, exploiting the 72$\times$ bandwidth hierarchy in modern GPU clusters, can reduce KV cache transfer latency by 3 to 18$\times$.
Co-optimizing MoE expert dispatch with KV cache placement and modeling CXL~3.0 as an overflow tier address complementary aspects of the problem.
While full evaluation awaits access to multi-node GPU clusters with heterogeneous interconnects, the complete system design and analytical models grounded in published specifications provide evidence that the approach merits further investigation and hardware validation.


\end{document}